%% file: tedloc_pr_published.tex
\documentclass[final,5p,times,twocolumn,10pt]{elsarticle}

\newcommand{\PRjournal}{Pattern Recognition}
\newcommand{\PRvolume}{183}
\newcommand{\PRyear}{2027}
\newcommand{\PRarticle}{114994}
\newcommand{\PRdoi}{10.1016/j.patcog.2026.114994}
\newcommand{\PRcitation}{\PRjournal\ \PRvolume\ (\PRyear) \PRarticle}
\newcommand{\PRrunningauthors}{S. Murtaza et al.}

\journal{\PRjournal}

\usepackage{enumitem}
\usepackage{amssymb}
\usepackage{amsmath}
\usepackage{booktabs}
\usepackage[table]{xcolor}
\usepackage{graphicx}
\usepackage{flushend}
\usepackage{subcaption}
\usepackage{xspace}
\usepackage{caption}
\usepackage{fancyhdr}
\usepackage{atbegshi}
\AtBeginShipout{\ifnum\value{page}=1 \global\hoffset=14pt\else\global\hoffset=0pt\fi}

\usepackage[T1]{fontenc}

\DeclareFontFamily{T1}{lmtt}{}
\DeclareFontShape{T1}{lmtt}{m}{n}{<->ssub*txtt/m/n}{}
\DeclareFontShape{T1}{lmtt}{m}{it}{<->ssub*txtt/m/it}{}
\DeclareFontShape{T1}{lmtt}{b}{n}{<->ssub*txtt/m/n}{}
\DeclareFontShape{T1}{lmtt}{bx}{n}{<->ssub*txtt/m/n}{}
\begingroup\fontencoding{T1}\fontfamily{txtt}\selectfont\endgroup
\DeclareFontShape{T1}{txtt}{b}{n}{<->ssub*txtt/m/n}{}
\DeclareFontShape{T1}{txtt}{bx}{n}{<->ssub*txtt/m/n}{}
\DeclareFontShape{T1}{txtt}{b}{it}{<->ssub*txtt/m/it}{}
\DeclareFontShape{T1}{txtt}{bx}{it}{<->ssub*txtt/m/it}{}

\makeatletter
\renewcommand\normalsize{%
  \@setfontsize\normalsize{8.2}{10.5}%
  \abovedisplayskip 8\p@ \@plus2\p@ \@minus4\p@
  \belowdisplayskip \abovedisplayskip
  \abovedisplayshortskip \z@ \@plus3\p@
  \belowdisplayshortskip 5\p@ \@plus3\p@ \@minus3\p@
  \let\@listi\@listI}
\renewcommand\small{\@setfontsize\small{7.7}{9.5}}
\renewcommand\footnotesize{\@setfontsize\footnotesize{7.2}{9}}
\renewcommand\scriptsize{\@setfontsize\scriptsize{6.5}{8}}
\renewcommand\large{\@setfontsize\large{11}{13}}
\renewcommand\Large{\@setfontsize\Large{13}{16}}
\renewcommand\LARGE{\@setfontsize\LARGE{15}{18}}
\makeatother

\definecolor{elslink}{RGB}{0,90,180}
\definecolor{elshead}{RGB}{0,90,180}
\definecolor{iccvblue}{RGB}{0,90,180}
\usepackage[breaklinks,colorlinks,allcolors=elslink]{hyperref}
\input{macros}
\renewcommand{\fakeslant}[2][0.25em]{#2}
\definecolor{lightgrayrow}{gray}{0.92}

\DeclareCaptionLabelSeparator{elsnewline}{\\}
\newcommand{\tabfont}{\footnotesize}

\fancypagestyle{elsheadings}{%
  \fancyhf{}%
  \fancyhead[L]{\footnotesize\itshape\PRrunningauthors}%
  \fancyhead[R]{\footnotesize\itshape\PRcitation}%
  \fancyfoot[C]{\footnotesize\thepage}%
}
\makeatletter
\def\ps@pprintTitle{%
  \let\@oddhead\@empty
  \let\@evenhead\@empty
  \def\@oddfoot{\hbox to \textwidth{\hfil\footnotesize\thepage\hfil}}%
  \let\@evenfoot\@oddfoot}

\newcommand{\PRpublishedBanner}{%
  \noindent\begingroup
  \setlength{\fboxsep}{0pt}%
  \colorbox{black!6}{\parbox[c][66pt][c]{\textwidth}{%
    \centering
    {\small This article was published in}\\[3pt]
    {\LARGE\bfseries\PRjournal}\\[5pt]
    {\small Volume \PRvolume\ (\PRyear), Article \PRarticle\ \textbar\ %
     \href{https://doi.org/\PRdoi}{https://doi.org/\PRdoi}}}}%
  \endgroup\par
  \vspace{12pt}}

\long\def\MaketitleBox{%
  \resetTitleCounters
  \def\baselinestretch{1}%
  \PRpublishedBanner
  \begin{flushleft}%
   \def\baselinestretch{1}%
    \Large\@title\par\vskip7pt
    \normalsize\elsauthors\par\vskip4pt
    \footnotesize\itshape\elsaddress\par\vskip8pt
    \hrule\vskip8pt
    \PRinfoabstract
    \vskip8pt\hrule\vskip8pt
  \end{flushleft}%
}

\newcommand{\PRinfoabstract}{%
  \noindent
  \begin{minipage}[t]{0.27\textwidth}
    \footnotesize\normalfont
    {\PRletterspace{A R T I C L E\quad I N F O}}\par\vskip6pt\hrule\vskip6pt
    \textit{Keywords:}\par
    \PRkeywordlist
  \end{minipage}\hfill
  \begin{minipage}[t]{0.69\textwidth}
    \footnotesize\normalfont
    {\PRletterspace{A B S T R A C T}}\par\vskip6pt\hrule\vskip6pt
    \small\PRabstracttext
  \end{minipage}\par}
\newcommand{\PRletterspace}[1]{\textrm{#1}}

\makeatletter
\newcommand{\PRbreak}{\ifhmode\vadjust{\penalty-\@M}\else\penalty-\@M\fi}
\newcommand{\PRnewpage}{\ifhmode\vadjust{\vfil\penalty-\@M}\else\vfil\penalty-\@M\fi}
\makeatother

\makeatother

\makeatletter
\gdef\emailauthor#1#2{\stepcounter{ead}%
     \g@addto@macro\@elseads{\raggedright%
      \let\corref\@gobble\eadsep{\href{mailto:#1}{#1}}
      (\PRemailname)\def\eadsep{\unskip,\space}}}
\newcommand{\PRemailname}{S. Murtaza}
\def\printFirstPageNotes{%
  \ifx\@tnotes\@empty\else\@tnotes\fi
  \ifx\@nonumnotes\@empty\else\@nonumnotes\fi
  \ifx\@cornotes\@empty\else\@cornotes\fi
  \ifx\@elseads\@empty\relax\else
   \let\thefootnote\relax
   \footnotetext{\textit{E-mail address:\space}\@elseads}\fi
  \ifx\@fnotes\@empty\else\@fnotes\fi}
\makeatother

\begin{document}

\newcommand{\PRabstracttext}{%
Weakly supervised object localization (WSOL) models can predict both the object class and the spatial regions corresponding to the object, without requiring explicit bounding-box annotations. Given their reliance on classification objectives, traditional WSOL methods, like class activation mapping, tend to focus on the most discriminative object regions, often missing the full spatial extent. Although vision-language models like CLIP encode rich semantic priors, their global text and class-token embeddings are not explicitly aligned with local patch embeddings, limiting patch-level localization. Recent methods such as GenPrompt address this limitation, but at the cost of increased complexity, as they rely on conditional denoising and elaborate prompt-learning strategies. In this paper, we propose Text Distillation for Localization (\tedloc), which distills knowledge from CLIP text embeddings to patch embeddings through contrastive alignment, thereby enabling patch-level foreground/background localization. A localization-guided classification module is also introduced, which uses localization scores to aggregate foreground patch embeddings for joint classification and localization within a single model. In addition, a QR-based orthogonalization of class text embeddings is applied before distillation to improve discrimination for semantically similar classes. Extensive experiments\textsuperscript{1} show that \tedloc improves \topone \loc by $\sim$5\% on \cubs and \ilsvrc, and \pxap by $\sim$31\% on histopathology benchmarks, while achieving more efficient inference than GenPrompt.}
\newcommand{\PRkeywordlist}{%
Deep Learning\\
Weakly Supervised Object Localization\\
Self-Supervised Learning\\
Vision Language Models\\
Knowledge Distillation.}


\begin{frontmatter}

\title{\tedloc: Text Distillation for Weakly Supervised Object Localization}

\author{Shakeeb Murtaza, Soufiane Belharbi, Alexis Guichemerre, Marco Pedersoli \& Eric Granger}
\address{Laboratoire d'imagerie, de vision et d'intelligence artificielle (LIVIA) \\ Dept. of Systems Engineering, ETS Montreal, Canada\\
  {\tt\footnotesize \textcolor{black}{\{shakeeb.murtaza.1, soufiane.belharbi.1, alexis.guichemerre.1\}@ens.etsmtl.ca}, \\ \{marco.pedersoli, eric.granger\}@etsmtl.ca }
}
\fntext[fn1]{Our code is available at{:} \href{https://github.com/shakeebmurtaza/TeDLOC}{github.com/shakeebmurtaza/TeDLOC}.}

\end{frontmatter}

\section{Introduction}

WSOL is a critical yet challenging task in computer vision, aiming to localize objects within images using a model trained using only image-class labels rather than instance-level annotations. The popular class activation mapping (CAM) method~\cite{zhou2016learning} leverages classification models for generating localization maps. However, they inherently focus on the most salient regions of an object and often fail to capture the full spatial extent~\cite{belharbi2022fcam}. This limitation arises because discriminative models are optimized to minimize mutual information between different instances of the same class. Various strategies have been proposed to mitigate this issue, including spatial regularization~\cite{wu2022background, xue2019danet}, adversarial erasing~\cite{zhang2018adversarial, choe2019attention, CHOE2021107949}, and leveraging pseudo-labeling~\cite{belharbi2022fcam,zhang2018self,murtaza2023discriminative}. However, these approaches are constrained by the local receptive fields of convolutional neural networks (CNNs), which limit their ability to capture global dependencies essential for complete object localization.

Vision transformers (ViTs)~\cite{gao2021ts} have recently shown potential in modeling long-range dependencies through self-attention. This addresses key limitations of CNNs for WSOL -- the restricted local receptive fields that prevent capturing global spatial dependencies, and the partial activation issue, whereby only the most discriminative object regions are highlighted~\cite{belharbi2022fcam}. However, they lack the local inductive biases\PRnewpage{} inherent to CNNs, often resulting in weaker local feature representations. Vision-language models, particularly contrastive language-image pre-training (CLIP), have emerged as a promising direction by aligning textual and visual features, which can be leveraged for localization tasks using class-level labels~\cite{lin23}. Yet, an architectural limitation of transformer-based models is the disconnection between the text embeddings and the local patch embeddings. Since the class token aggregates global image semantics rather than spatially grounded representations, it remains poorly aligned with the local features required for localization, making it challenging to extract accurate localization maps. Moreover, the predominant approaches for extracting localization maps from CLIP, such as gradient-based and attention manipulation methods, rely heavily on ground truth (GT) class information. This dependency leads to performance degradation when predicted classes derived from an external model are employed because of feature misalignment and class confusion, as illustrated by CLIP’s frequent conflation of similar classes like ``airplane'' and ``aircraft''~\cite{wang2024learn}. Consequently, employing a CLIP model without fine-tuning introduces substantial errors in downstream tasks, prompting the need for strategies that can learn precise localization cues while minimizing reliance on explicit class labels.\PRnewpage{}

\begin{figure}[!t]
     \centering
    \begin{subfigure}[t]{\linewidth}
        \centering
        \includegraphics[width=\linewidth]{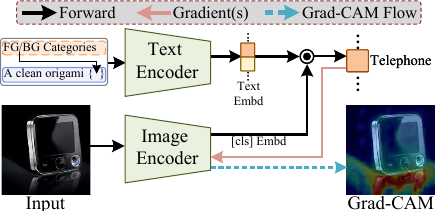}
        \caption{CLIP-ES localization via standard Grad-CAM~\cite{lin23} at test time.}
        \label{finalch:fig:main-clip}
    \end{subfigure}
    \vspace{0.2cm}

    \begin{subfigure}[t]{\linewidth}
        \centering
        \includegraphics[width=\linewidth]{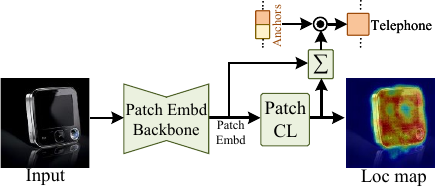}
        \caption{\tedloc localizes an object without class labels at test time.}
        \label{finalch:fig:main-simloc}
    \end{subfigure}
\caption{A comparison of our \tedloc versus CLIP-ES~\cite{lin23} methods for extracting localization maps from CLIP at test time. \textbf{(A)} CLIP-ES utilizes Grad-CAM to extract localization maps from CLIP, requiring GT class labels during inference. \textbf{(B)} In contrast, our \tedloc model distills knowledge from CLIP text embeddings into the visual encoder during training, allowing it to produce both classification scores and localization maps without requiring class labels during inference.}\label{finalch:wacveval:fig:issue-model-selection}
\end{figure}

Recently, GenPrompt~\cite{zhao2023generative} attempted to address these challenges by framing WSOL as a conditional denoising process, leveraging CLIP embeddings to capture discriminative regions. While GenPrompt improves localization by using CLIP embeddings, it still relies on external classifiers or GT class labels during inference, adding to their complexity and limiting their applicability in real-world scenarios. Moreover, despite robust map generation capabilities, a fundamental limitation remains the inability of CLIP-based methods to localize objects within an image without prior class information. This constraint poses a significant challenge for downstream tasks, as they require computing class labels beforehand (Fig.\ref{finalch:fig:main-clip}). Given these challenges, we seek to effectively harness vision-language models to learn precise localization cues for WSOL while mitigating misclassification and reducing reliance on GT labels during inference.

To address these challenges, we propose Text Distillation for Localization (\tedloc). CLIP's text embeddings encode semantic information that aligns strongly with visual concepts. However, CLIP cannot align text embeddings with local patch-level representations, limiting its ability to localize objects. \tedloc addresses this gap by distilling knowledge from CLIP's text embeddings, which serve as a powerful link between visual and textual modalities.

Localization information is learned by transferring knowledge from text embeddings to the localization module (Fig.\ref{finalch:fig:main-simloc}). Using contrastive learning within a teacher-student framework, the patch-level visual representations of our model are aligned with text embeddings. This alignment is guided by pseudo-labels that can be extracted using an off-the-shelf method based on CAMs. Learning from text embeddings allows \tedloc to achieve state-of-the-art performance using one model that is selected using the best localization performance without requiring separate classifiers trained and selected using the validation set over classifier scores. \tedloc introduces a new paradigm where\PRbreak{} classification is achieved through localization, thereby eliminating the need for model selection over classifier scores. By training the localization module to distinguish between foreground (FG) and background (BG) regions based on their similarity to text embeddings, \tedloc enables the model to classify and localize objects simultaneously. Furthermore, to address the limitations of CLIP's frequent conflation of semantically similar classes, we propose a method to orthogonalize text embeddings before distillation. By default, text embeddings in CLIP may not be sufficiently discriminative between similar classes because of their proximity in the embedding space. To mitigate this issue, we decompose the embeddings using QR decomposition~\cite{gander1980algorithms} and utilize the resulting orthogonal basis vectors for alignment.

Our \tedloc method employs a transformer-based architecture that decomposes an image into a set of patches, generating upsampled patch embeddings through our model backbone to produce a fine-grained localization map. Each patch embedding is individually classified to estimate its likelihood of representing a FG or BG region. These classification scores are then gathered to produce a localization map, highlighting regions of interest within the image. The global classification score for the entire image is a weighted average of the patch embeddings. Our approach is inspired by the multiple instance learning (MIL) framework~\cite{carbonneau2018multiple}, where each image consists of a  ``bag'' containing multiple ``instances'' (image patches) with only bag-level labels available during training. Leveraging MIL, object localization and classification are performed simultaneously by assigning higher weights to discriminant patches, without relying on external classifiers or prior class information. This aligns well with WSOL, enabling our model to independently produce accurate localization maps and classification scores.

\noindent \textbf{Our main contributions are summarized as follows.}\\
    \noindent \textbf{(1)} The \tedloc method is introduced that distills knowledge from CLIP text embeddings into a patch embedding using contrastive learning. It allows for patch-level foreground/background localization without requiring class labels or external classifiers at inference time.
    \\
    \noindent \textbf{(2)} A classification module is introduced that leverages localization scores to compute the expected embeddings of FG regions. Using a weighted average of patch embeddings, where the weights are derived from the FG localization map. This ensures that FG embeddings align with the correct class embeddings. This eliminates the need for an external classifier and allows our model to classify and localize simultaneously.
    \\
    \noindent \textbf{(3)} To mitigate the tendency of CLIP to confuse semantically similar classes, a QR-based orthogonalization of class text embeddings is proposed before distillation, improving discriminability for both localization and classification.\\
    \noindent \textbf{(4)} An extensive set of experiments on natural (\cubs and \ilsvrc) and histology (\glas and \camelyonsev) image datasets indicate that \tedloc can outperform state-of-the-art WSOL methods on several challenging benchmarks.

\section{Related Work}

\noindent \textbf{(a) Weakly supervised object localization.}
WSOL is a challenging task that seeks to localize objects using only image-level supervision. The foundational work in WSOL ~\cite{zhou2016learning} proposes to harvest CAMs from pre-trained CNNs, leveraging global average pooling (GAP) to guide the network’s attention toward specific regions in an image. Despite its impact, CAM and related CNN-based approaches are constrained to highlight discriminative regions, often neglecting complete object extents. This limitation has led to the development of different WSOL methods that can look beyond discriminative regions. Furthermore, \cite{HUI2022108664} proposes to utilize inter-class and intra-class gradients to improve CAMs.\PRbreak{}

Erasing-based methods aim to mitigate CAM partial activation by selectively obscuring parts of an image to encourage broader localization. HaS~\cite{singh2017hide} and CutMix~\cite{yun2019cutmix} employ random erasure, which forces the network to explore different object parts beyond discriminative regions. Building on this, adversarial erasing methods like ACoL~\cite{zhang2018adversarial} and ADL~\cite{choe2019attention} use dual classifiers to identify and erase dominant regions, uncovering complementary object regions. Techniques like SPG~\cite{zhang2018self} go further, integrating pixel-wise correlation constraints to maintain context and consistency across object regions.

Other works target the inherent challenge in CNNs to capture only local semantic features due to limited receptive fields. Consequently, newer methods leverage structural cues and integrate BG suppression techniques. SPA~\cite{pan2021unveiling} enhances structural consistency, while PSOL~\cite{zhang2020rethinking} introduces a two-stage WSOL approach that decouples classification from localization tasks, providing robust pseudo-annotations for regression without class constraints. Methods such as BAS~\cite{wu2022background} reinforce this separation by suppressing BG regions and emphasizing FG areas critical to localization.

To overcome the inherent CAM limitations in capturing long-range dependencies, transformer-based approaches for WSOL are gaining traction. Transformers, known for their self-attention mechanism, enable networks to capture both local and global feature dependencies.
{More recently, CDTR~\cite{chen2025clip} proposes embedding learnable category biases into the self-attention maps of a vision transformer (ViT) to learn category-aware representations for accurate localization.}
In WSOL, TS-CAM~\cite{gao2021ts} leverages token-patch fusion with semantic maps to improve spatial coherence. By exploiting the long-range capability of transformers, these methods significantly broaden the object localization scope and address the core issues of CNN-based localization methods. Extending WSOL to video, \cite{BELHARBI2025111358} proposes exploiting color cues across frames to improve localization maps. Beyond closed-set settings, \cite{XIE2026111808} introduces open-world WSOL for unseen categories via contrastive representation co-learning.

\noindent \textbf{(b) Contrastive language–image pre-training.} CLIP~\cite{radford2021learning} is a foundational model designed to align visual and language representations, trained on 400 million image-text pairs collected from web data. By learning from paired data, CLIP produces a robust model capable of zero-shot adaptation to diverse tasks by computing similarities between images and textual descriptions. In WSOL, GenPrompt~\citep{zhao2023generative} leverages CLIP to identify discriminative regions and employs VQGAN for embedding generation within a denoising process to localize objects; however, its computational complexity hinders real-time applicability.

Moreover, CLIP has been widely adapted for weakly supervised semantic segmentation (WSSS), prompting multiple approaches to leverage its capabilities for generating class activation maps (CAMs) without extensive annotation. For instance, CLIMS~\cite{xie2022clims} uses CLIP to enhance the completeness of object regions within CAMs while suppressing BG regions. CLIP-ES~\cite{lin23} employs GradCAM to extract CAMs directly from CLIP, demonstrating that activations can be generated without fine-tuning. SCLIP~\cite{wang2023sclip} and NACLIP~\cite{sina25wacv} propose modifying the last attention block to produce score maps for each patch embedding, enabling segmentation map generation without requiring a backward pass.
{Beyond attention manipulation, WeakCLIP~\cite{zhu2025weakclip} introduces a text-to-pixel matching paradigm for WSSS. It exploits CLIP knowledge to refine CAMs.}
Similarly, \cite{yang2024foundation} introduces learnable prompts in CLIP and uses pseudo-labels from SAM~\cite{kirillov2023segment} for fine-tuning for WSSS.

While CLIP-based methods yield competitive maps, these methods rely on hand-crafted textual templates and predefined class representations, such as prompts like ``a photo of [CLS].'' This requires prior knowledge of the specific class name before producing a localization map of each image. This reliance restricts the model's adaptability across different computer vision tasks. It also leads to a substantial decline in performance when using predicted classes due to feature misalignment and class confusion. Furthermore, the disconnection between CLIP's text embeddings and its local patch embeddings limits\PRbreak{} the model's ability to capture spatially grounded representations. These limitations underscore a critical challenge: the need for WSOL methods that can learn precise localization cues without reliance on explicit class labels or predefined textual templates. To address this challenge, we propose a novel method that distills knowledge from text-image representations of CLIP to guide the localization network. Additionally, we mitigate the tendency of CLIP to conflate semantically similar classes by orthogonalizing the text embeddings before alignment, reducing semantic overlap and improving discriminability. This enables the model to classify the image by computing the similarity between visual embeddings and class anchors as visual embeddings are pushed toward class-text embeddings.

\section{The Proposed \tedloc Method}
\label{sec:tedloc}

Let us consider a training set $\mathbb{D} = \{(x_i, y_i)\}_{i=1}^{N}$ of $N$ images, where each image $x_i \in \mathbb{R}^{H \times W \times 3}$ is associated with an image-level label $y_i \in \{1, \dots, K\}$, representing one of $K$ object classes, with no bounding box (bbox) supervision. WSOL methods seek to train a model for object localization and classification using only image-level labels. In this paper, we leverage text embeddings from a pre-trained vision-language model, specifically CLIP~\cite{sina25wacv}.
\begin{figure*}[!t]
    \centering
    \includegraphics[width=0.60\linewidth]{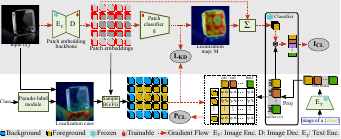}
    \caption{An overview of the \tedloc method for distilling FG text embeddings into the patch embedding backbone. First, pseudo-labels are extracted to guide the identification of FG and BG patches. By leveraging these FG/BG regions, the model minimizes the similarity of $E_V$ with the relevant text embedding for FG classes while maximizing dissimilarity with embeddings of other classes. Using a binary FG/BG classifier, \tedloc generates localization maps by classifying patches as FG or BG and producing class probabilities for image classification. This joint task enables the model to produce both accurate localization and classification outputs without explicit bounding box supervision.}
    \label{finalch:fig:system-diagram}
\end{figure*}

Our model (see Fig.\ref{finalch:fig:system-diagram}) consists of a patch embedding backbone network and a compact head for localization and classification tasks. The backbone network is comprised of the (i) \textit{Encoder \(E\)}, a ViT (ViT-EVA-L)~\cite{fang2023eva} that decomposes images into patches and produces patch-level embeddings. It is pre-trained for classification and frozen during training. (ii) The \textit{Decoder \(D\)} upscales these patch embeddings to a high spatial resolution where the $p^{\text{th}}$ output patch is denoted as $z_p \in \mathbb{R}^d$. The CLIP text encoder is denoted as $E_T$. Moreover, we introduce a binary patch classifier $g(z_p)$ that predicts the FG/BG for each patch. Its response across all patches forms a localization map $M \in [0,1]^{H \times W}$ that contains the FG object associated with the image class. This map provides the localization generated by our method. Furthermore, we define the classification function  {${f: \mathbb{R}^d \to [0,1]^K}$}, which is parametrized with \emph{frozen} class weight vectors ${t_k \in \mathbb{R}^d}$ for a class $k$ for ${k \in \{1, \dots, K\}}$. {Given an embedding vector ${v \in \mathbb{R}^d}$, the function $f$ returns a probability distribution over the $K$ classes by applying a softmax to the dot-product similarities between $v$ and all class anchors, ${f(v)=\mathrm{softmax}\!\big(\langle v, t_1 \rangle, \dots, \langle v, t_K \rangle\big)}$. Therefore, its $k^{\text{th}}$ entry ${f_k(v)}$ is the predicted probability of class $k$.} In this work, we consider ${t_k}$ as the CLIP text embedding of the class $k$, while $v$ could be a patch embedding $z_p$, or the global image embedding $h$ of our method.

\subsection{Key Components}

The rest of this subsection introduces two important components of \tedloc: the generation of pseudo-labels of patches and the pre-processing of text embeddings.

\noindent \textbf{Patch-level pseudo-label generation.} To train our model, we propose to leverage patch-level pseudo-annotations corresponding to FG and BG patches. However, since such annotations are not available in the WSOL setting, we consider an off-the-shelf pre-trained classification model with a CAM method. Such models can yield a discriminative map to localize a target class, which is adequate during training. Generally, any CAM-based model can be used~\cite{choe2020evaluating, murtaza2025realistic, rony2023deep} to generate localization cues that can be leveraged for training. During the training of our model, we randomly sample a few FG/BG patch locations for each training image at each training step to avoid overfitting. This pseudo-labeling technique has been effective in guiding WSOL learning~\cite{belharbi2022fcam, murtaza2023dips}. In our experiments, we sample the same number of FG/BG locations to maintain a balanced ratio between the two classes. $\omega$ denotes the set containing the sampled patches for both FG/BG, while $\omega^+$ contains only\PRbreak{} the selected FG patches. $y^{\prime}_p \in \{0,1\}$ is used as the patch pseudo-label where $0$ is BG and $1$ is FG.

\noindent \textbf{Sampling FG/BG regions for pseudo-label generation.} To train our model, we employ pseudo-labels for FG and BG regions following recent methods~\cite{belharbi2022fcam, murtaza2023discriminative, murtaza2023dips}. We obtain these pseudo-labels by utilizing CAM \(C \in \mathbb{R}^{H \times W}\) extracted from a pre-trained classifier model. These CAMs indicate regions of the image highlighting an object belonging to the FG class, which can be used to guide the sampling of FG and BG regions. We first apply Otsu's thresholding method~\cite{otsuthresh79} to the CAM \(C\) to separate FG and BG regions. This method automatically determines a threshold, effectively separating high-activation regions (FG) from low-activation ones (BG). We denote the set of pixel locations in the image domain as \(\Omega\).

For FG sampling, we focus on regions with high activation values in \(C\). Specifically, we select the top \(n^+\) pixels with the highest activation values inside the image, forming the set of potential FG locations \(\omega_{\text{+}} \subset \Omega\). We then randomly sample a subset of these locations to be used as FG samples during training. For BG sampling, we consider regions with low activation values in \(C\). We select the bottom \(n^-\) pixels with the lowest activation values, excluding any pixels that are within the FG regions. This forms the set of potential BG locations \(\omega_{\text{-}} \subset \Omega\). We randomly sample from these locations to obtain BG samples for training. We note that \(\omega = \omega_{\text{+}} \cup \omega_{\text{-}}\) is the set of all sampled pixels in both FG and BG in one sampling step.

To ensure that our model generalizes well and avoids overfitting, we perform this sampling process at every training step for each image. This dynamic sampling allows the model to explore different regions of the image during training, promoting robustness and consistency in learning.

The sampled FG and BG pixel locations are used to create a partial pseudo-label mask \(y^\prime_p \in \{0, 1\}\), where \(y^\prime_p = 1\) for FG pixels, \(y^\prime_p = 0\) for BG pixels, and locations with unknown labels are left undefined. The set of sampled locations $p$ is defined by \(p \in \omega\).

\begin{figure*}[!t]
    \centering
    \includegraphics[width=0.74\linewidth]{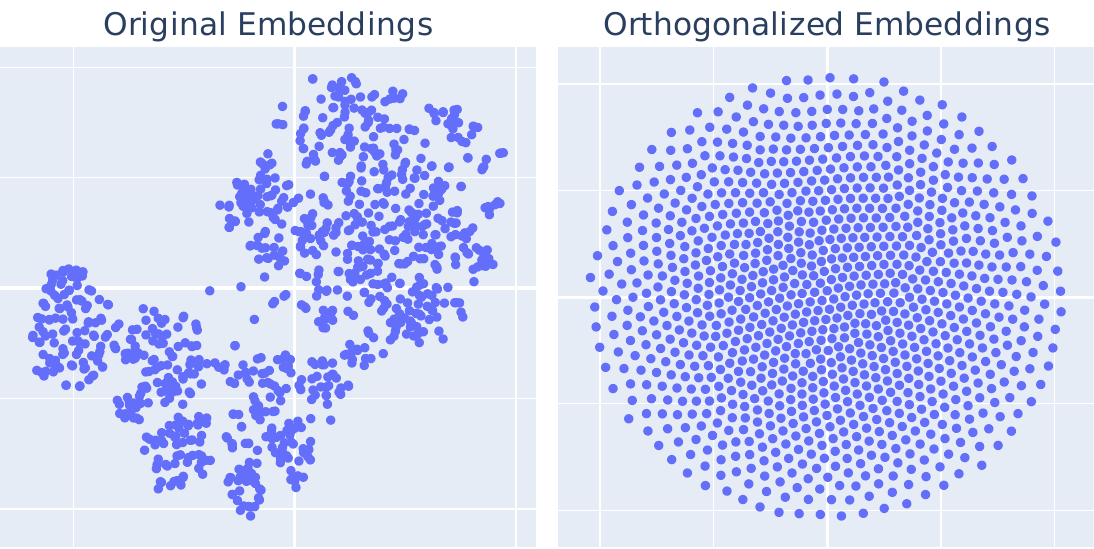}
    \caption{t-SNE visualizations of CLIP text embeddings for ILSVRC~\cite{imagenet_cvpr09} classes before and after orthogonalization. \textbf{(Left)} Prior to orthogonalization, embeddings of semantically similar classes (e.g., ``airplane'' and ``aircraft'') cluster closely together, leading to potential confusion. \textbf{(Right)} After orthogonalization (QR decomposition), the embeddings are more uniformly distributed and orthogonal, reducing overlap.}
    \label{finalch:fig:orth-embds}
\end{figure*}

\noindent \textbf{Text embeddings orthogonalization.} CLIP text embeddings of classes are used to distill localization knowledge as they provide a powerful link between global visual and textual representations. However, text embeddings can sometimes conflate similar classes due to semantic overlap (e.g., ``airplane'' and ``aircraft'')~\cite{wang2024learn}. This overlap between classes limits the benefit of those embeddings, especially when used in discriminative scenarios. To mitigate this issue, we propose to pre-process the class-text embeddings before using them in our method. We consider a transformation that projects all the text embeddings into a space where the distance between each pair of embeddings is maximum. In this work, we use orthogonal projection, in particular, QR orthogonalization~\cite{gander1980algorithms} and conserve the basis of the projection. In\PRbreak{} the rest of this paper, the orthogonalized version of text embeddings for class $k$ is referred to as $t_k$. These new embeddings are kept frozen and play the role of class anchors that are carried in our model, allowing us to discard the text encoder. Fig.\ref{finalch:fig:orth-embds} illustrates the issue of class overlap and the impact of orthogonalization over text embeddings.

\vspace{0.2cm}
\subsection{\tedloc Training Method}

\noindent \textbf{Text embedding distillation to local patches.}
Our objective is to distill text-class embeddings into patch embeddings and to discard the text encoder. Since text and global image embeddings in CLIP are not directly tied to patch embeddings, text embeddings cannot directly be used to localize objects. Patch embeddings are not necessarily correlated with text embeddings~\cite{sina25wacv}. In this work, we propose to create a direct link between text embeddings and patch embeddings for semantic localization. To achieve this, we propose to use knowledge distillation, where we transfer the text embedding of the image class to the patch embeddings, allowing us to perform object localization. In particular, a contrastive learning loss~\cite{radford2021learning} is employed to ensure that FG patch embeddings are similar to the text embedding of the image class while being dissimilar from embeddings of other classes. To this end,
{only the embeddings $z_p$ of FG patches, with $p \in \omega^+$, are used for this loss.} It can be defined through a standard cross-entropy as follows~\cite{radford2021learning},
\begin{equation}\label{finalch:eq:kd}
\mathbf{L}_{\text{KD}} = \sum_{p \in \omega^+} CE(y, {f(z_p)}).
\end{equation}
where $t_y \in \mathbb{R}^d$ denotes the orthogonalized text embedding of the GT class $y$, that is, the class anchor for the image. Concretely, for each foreground patch {embedding} $z_p$, {$f(z_p)$ is the class distribution obtained by a softmax over the dot-product similarities $\langle z_p, t_k \rangle$ between $z_p$ and \emph{all} $K$ class text anchors}. The cross-entropy in Eq.~(1) therefore trains $z_p$ to maximize its dot-product similarity to the ground-truth anchor $t_y$ while simultaneously minimizing similarity to every other anchor $t_k,\ k \neq y$; the correct class acts as the single positive and the remaining $K{-}1$ class embeddings act as negatives, which is precisely the one-vs-all contrastive objective used in CLIP~\cite{radford2021learning}. Furthermore, Eq.\ref{finalch:eq:kd} is computed only on the selected FG patches $\omega^+$, as BG patches lack corresponding class embeddings, rendering them unsuitable for this contrastive loss. Since BG is not considered, this can lead to poor localization as the BG region is present in most images. To mitigate this issue, we introduce a patch binary FG/BG classifier $g$, which repels BG patch embeddings from FG text embeddings.
{It is trained using the\PRbreak{} embeddings $z_p$ of both FG and BG patches and their pseudo-labels $y^{\prime}_p$}
via standard cross-entropy loss,
\begin{equation}\label{finalch:eq:patchcls}
\mathbf{P}_{\text{CL}} = \sum_{p \in \omega} CE(y^{\prime}_p, g(\mathbf{z}_p)).
\end{equation}
Minimizing this loss allows for both FG and BG regions to be present in the image, which helps avoid imbalanced localization.

\noindent \textbf{Global image embeddings from local patch embeddings for classification.} So far, our method can only perform localization. To further allow it to perform image classification, our aligned FG patch embeddings are leveraged to construct a global image embedding that describes the object in the image. This creates a reversed link from local patch representations to global image representation, allowing us to learn to classify the image. This aligns perfectly with our distillation approach from class text embedding to patch embeddings described previously, where we ensure that FG patch embeddings are correlated with the text embedding of the image class. Therefore, we leverage this property to construct a global image embedding $h$ using all the patch embeddings and the patch classifier $g$ as follows,
\begin{align}\label{finalch:eq:globalembd}
h&=\sum_p{a_p}z_p\quad,
\text{where } a_p=g({z_p})/\sum_j{g(z_j)}.
\end{align}

Eq.\ref{finalch:eq:globalembd} performs a weighted average of the embeddings for all the patches by giving more importance to patches that are classified as FG since their $g(z_p)$ will be close to $1$. In addition, BG patches are discarded since $g(z_p)$ is close to $0$. This effectively performs a differentiable selection of FG patches, allowing for training with gradient-based methods. Most importantly, the final aggregated embedding $h \in \mathbb{R}^d$ is expected to resemble the text embedding of the image class $t_k$. To furthermore ensure this, this embedding is trained to be as close as possible to $t_k$ using standard cross-entropy as follows,
\begin{equation}\label{finalch:eq:imagecls}
\mathbf{I}_{\text{CL}} = CE(y, {f(h)}).
\end{equation}

\noindent\textbf{Overall training loss.} Our overall training loss contains the three terms discussed previously: knowledge-distillation loss ($\mathbf{L}_{\text{KD}}$), patch classifier loss ($\mathbf{P}_{\text{CL}}$) and global image classification loss ($\mathbf{I}_{\text{CL}}$) as follows,
\begin{equation}\label{finalch:eq:overall-loss}
{\mathbf{L} = \lambda_{1} \mathbf{L}_{\text{KD}}+ \lambda_{2}  \mathbf{P}_{\text{CL}} + \lambda_{3} \mathbf{I}_{\text{CL}}} \ ,
\end{equation}
where \(\lambda_1\), \(\lambda_2\) and \(\lambda_3\) are weighting factors that balance the contribution of each term. Stochastic Gradient Descent (SGD) is used for optimizing the parameters of our model (parameters of the decoder $D_I$ and patch classifier $g$). By jointly optimizing these loss functions, our model learns to produce discriminative and well-aligned visual representations, enabling simultaneous classification and localization.\PRbreak{}

\vspace{0.25cm}
\section{Results and Discussion}

\subsection{Experimental Methodology}

\noindent \textbf{Dataset.} \tedloc is evaluated on challenging natural and histology image benchmarks to assess its performance across standard WSOL scenarios.

\emph{Natural images:} Two common challenging datasets were used for our WSOL experiments. (i) Caltech-UCSD birds-200-2011 (\cubs) \cite{wah2011caltech} consists of 11,788 images spanning 200 bird species. The dataset is partitioned into 5,994 training images and 5,794 testing images. For validation, an independent set of 1,000 images (five per class) collected by~\cite{choe2020evaluating} is utilized. (ii) ImageNet large-scale visual recognition challenge (\ilsvrc)~\cite{imagenet_cvpr09} includes approximately 1.2 million training images and 10,000 validation images across 1,000 classes. We use the original validation split as our test set due to its sufficient size for robust evaluation. For validation purposes, \ilsvrcvtwo, collected by \cite{recht2019imagenet} and annotated by~\cite{choe2020evaluating}, is employed to mitigate biases toward the test set. For a fair comparison, we strictly adhere to the commonly used WSOL protocol proposed in~\cite{choe2020evaluating} for both datasets.

\emph{Histology images:} We evaluate \tedloc on two histopathology benchmarks covering distinct clinical settings: \glas for colon cancer and \camelyonsev for breast cancer. \glas provides colon histology images with gland-level annotations~\cite{rony2023deep} and contains 67 training images, 18 validation images for classification, 6 validation images for localization, and 80 test images. For breast cancer, we use a recent WSOL benchmark derived from \camelyonsev~\cite{guichemerre2026adaptwsol}, in which the original whole-slide dataset is reformulated into a patch-level setting suitable for weakly supervised classification and localization. Following~\cite{guichemerre2026adaptwsol}, for each center and each class, 8/2/2 whole-slide images are selected for training, validation, and testing, respectively. Tumor and normal patches are then extracted using the provided tumor masks, while only informative tissue regions are retained to avoid trivial patches, as defined in~\cite{rony2023deep, guichemerre2026adaptwsol}. The numbers of training, test, and validation images used for evaluation for each \camelyonsev center are summarized in Tab~\ref{tab:datasplit} as introduced by~\cite{guichemerre2026adaptwsol}.

\noindent \textbf{Evaluation measures.} Following earlier work by Choe et al. \cite{choe2020evaluating}, we employ three localization measures alongside a classification measure to evaluate the proposed and baseline methods. The localization measures are as follows. (1) \maxboxacc (referred to in previous work as \corloc \cite{deselaers2012weakly} or \gtknown \cite{singh2017hide}), which quantifies the proportion of images for which the predicted bbox achieves an Intersection over Union (\iou) threshold of $\sigma = 50\%$ with the ground-truth bbox, independent of classification accuracy (\cl); (2) Top-1 localization accuracy\PRbreak{}
\begin{table}[!t]
\caption{Dataset splits used in~\cite{guichemerre2026adaptwsol} for \camelyonsev.}
\label{tab:datasplit}
\tabfont
\setlength{\tabcolsep}{4pt}
\begin{tabular*}{\linewidth}{@{\extracolsep{\fill}}lcccccc@{}}
\toprule
 & \glas & \camelyonsevzero & \camelyonsevone & \camelyonsevtwo & \camelyonsevthree & \camelyonsevfour \\
\midrule
Train      & 67 & 634 & 1066 & 498 & 816 & 940 \\
Val (CL)   & 18 & 144 & 38 & 110 & 102 & 146 \\
Val (PxAP) & 6 & 10 & 10 & 10 & 10 & 10 \\
Test       & 80 & 262 & 172 & 448 & 376 & 298 \\
\bottomrule
\end{tabular*}
\end{table}
\begin{table*}[!t]
\caption{\maxboxacc, \topone\loc, and \topfive\loc performance of \tedloc against state-of-the-art methods on the ILSVRC and CUB datasets. The first row corresponds to Grad-CAM for CLIP~\citep{lin23}, where class labels are required to produce text embeddings for extracting localization maps. Noteably, \tedloc outperforms existing methods without relying on text-encoder. {
The row \tedloc* ($\langle$patch, anchor$\rangle$) reports a GT-known analysis variant, where localization maps are computed from patch--class similarity scores via the dot product $\langle z_p, t_y \rangle$. Here, $z_p$ denotes the patch embedding at location $p \in \Omega$, and $t_y$ is the text anchor associated with the ground-truth class $y$. This variant is used to assess the quality of the text-to-patch alignment.
}}
\label{finalch:tab:results_ilsvrc_cub}
\tabfont
\begin{tabular*}{\textwidth}{@{\extracolsep{\fill}}lcccccc@{}}
\toprule
\textbf{Method} & \multicolumn{3}{l}{\ilsvrcbold} & \multicolumn{3}{l}{\cubsbold}\\
\cmidrule{2-4}\cmidrule{5-7}
 & \textbf{\maxboxacc} & \toponebold \locbold & \topfivebold \locbold & \textbf{\maxboxacc} & \toponebold \locbold & \topfivebold \locbold  \\
\midrule
CLIP-ES (Pred)~\citep{lin23} {\emph{(cvpr,2023)}} & 71.2 & -- & -- & 91.6 & -- & -- \\
TS-CAM~\citep{gao2021ts} {\emph{(iccv,2021)}} & 67.7 & 53.4 & 64.3 & 71.3 & 83.8 & 87.7 \\
SCM~\citep{bai2022weakly} {\emph{(eccv,2022)}} & 68.8 & 56.1 & 66.4 & 76.4 & 91.6 & 96.6 \\
LCTR~\citep{ChenWWJSTWZC22} {\emph{(aaai,2022)}} & 68.7 & 56.1 & 65.8 & 92.4 & 79.2 & 89.9 \\
\midrule
PSOL~\citep{zhang2020rethinking} {\emph{(cvpr,2020)}} & 66.3 & 58.0 & 65.0 & 91.8 & 80.9 & 90.0 \\
C\textsuperscript{2}AM~\citep{xie2022c2am} {\emph{(cvpr,2022)}} & 68.5 & 59.6 & 67.1 & 92.9 & 81.8 & 91.1 \\
GenPrompt~\citep{zhao2023generative} {\emph{(iccv,2023)}} & 75.0 & 65.2 & 73.4 & 98.0 & 87.0 & 96.1 \\
CATR~\cite{chen2023category} {\emph{(iccv,2023)}} & 69.2 & 56.9 & 66.6 & 94.9 & 79.6 & 92.0 \\
DA-WSOL~\cite{zhu2024boosting} {\emph{(pami,2024)}} & 71.8 & 55.3 & -- & 88.4 & 71.1 & -- \\
BAS~\cite{wu2022background} {\emph{(ijcv,2022)}} & 72.0 &  58.5 & 69.0 &  94.6 & 72.0 & 88.1 \\
SeCM~\cite{CAO2025110971} {\emph{(pr,2025)}} & 72.9 &  61.7 & 70.9 &  98.0 & 81.9 & 94.2 \\
\midrule
\tedloc (ours) & \textbf{75.6} & \textbf{70.0} & \textbf{75.1} & \textbf{98.7} & \textbf{91.7} & \textbf{97.6}   \\
\tedloc\hspace{-0.15cm}* ($<$patch,anchor$>$) & \textbf{77.1} & - & - & \textbf{98.7} & - & -\\
\bottomrule
\end{tabular*}
\end{table*}
 (\topone \loc), measuring the proportion of images where the model’s top predicted class is correct and the bbox IoU with ground truth exceeds $\sigma = 50\%$; and (3) Top-5 localization accuracy (\topfive \loc), defined as the proportion of images for which the true class label is within the model's top-five predictions and the bbox meets an IoU of $\sigma = 50\%$. For histology, to evaluate localization performance, we use \pxap, a metric introduced in~\cite{choe2020evaluating} that is based on pixel-wise precision and recall and has been widely used in histology benchmarks~\cite{rony2023deep,guichemerre2026adaptwsol,guichemerre2024source}.

\noindent \textbf{Implementation details.} We closely followed the experimental setup of Choe et al. \cite{choe2020evaluating}, dataset splits, evaluation of localization maps across multiple thresholds, and training epochs. Specifically, 50 epochs for the \cubs dataset and 10 epochs for \ilsvrc. Furthermore, our model is trained with a batch size of 32 and 16 for \ilsvrc and \cubs, respectively. In Eq.\ref{finalch:eq:overall-loss}, the hyper-parameters $\lambda_1, \lambda_2 \text{ and } \lambda_3$ used in the total training loss (Eq.\ref{finalch:eq:overall-loss}) terms that are optimized over the values $(0, 1]$. Optimization of our model was performed using SGD, with a learning rate from 1e-6 up to 0.01. We also fine-tuned the weight decay and momentum. In our experiments, we use EVA-CLIP pre-trained on natural images together with CLIP-ES as the CAM module. Localization maps were evaluated at a resolution of $256\times256$.

For histology, the compared WSOL baseline methods are trained for 1000 epochs on \glas and 500 epochs on \camelyonsev. We use a weight decay of 0.0001. During training, images are first resized to $256\times256$ and then randomly cropped to $224\times224$, with the learning rate selected from \{0.0001, 0.001\} and the learning-rate decay factor fixed to 0.9. For \tedloc, we follow the same training setup as in the natural-image setting. Specifically, images are first resized to $512\times512$ and then randomly cropped to $448\times448$. Due to the size of the model, training is performed for 500 epochs on \glas and 200 epochs on \camelyonsev. Following prior work~\cite{belharbi2022fcam,guichemerre2024source}, we perform a hyper-parameter search over the loss weights \(\lambda_1\), \(\lambda_2\) and \(\lambda_3\) in \{1, 2\}, and over the learning rate in \{0.0001, 0.001\}. We use the pretrained CONCH vision-language model~\cite{conch}, including both its visual and text encoders. To ensure direct comparability, we adopt the same experimental protocol, evaluation metrics, and baseline methods as PixelCAM~\cite{belharbipixelcam}. Moreover, applying \tedloc to histology requires a slight modification of its original training strategy. In its initial form, \tedloc aligns foreground pixel embeddings with the text embedding of the corresponding class. For \camelyonsev, this assumption does not hold for the normal class, since normal patches do not contain foreground regions. We therefore modify the alignment strategy by associating background embeddings with the text embedding of the normal class. To avoid an excessive number of background embeddings relative to foreground ones, we follow the strategy used in PixelCAM~\cite{belharbipixelcam} and retain a proportion $\rho$ of background embeddings, with $\rho$ fixed to 0.5 in all experiments. For \glas, we use the original \tedloc setting, since both normal and cancer images contain foreground/background tissue for alignment.

\noindent \textbf{Baseline methods.} For natural-image benchmarks, we compare \tedloc with recent state-of-the-art WSOL methods, including TS-CAM~\citep{gao2021ts}, SCM~\citep{bai2022weakly}, LCTR~\citep{ChenWWJSTWZC22}, C$^2$AM~\citep{xie2022c2am}, PSOL~\citep{zhang2020rethinking}, DiPS~\cite{murtaza2023dips}, CATR~\cite{chen2023category}, DA-WSOL\cite{zhu2024boosting}, BAS~\cite{wu2022background}, and GenPrompt~\citep{zhao2023generative}. Additionally, CLIP-ES~\cite{lin23}, which utilizes Grad-CAM to extract localization maps from CLIP, was employed in a zero-shot setting. Specifically, we considered two variants of this method: CLIP-ES (GT-Known) and CLIP-ES\PRbreak{} (Pred). The CLIP-ES (GT-Known) variant requires ground-truth class labels during inference to generate localization maps. While this provides an upper bound on performance, it relies on privileged information unavailable in practical WSOL scenarios, thereby limiting its applicability. In contrast, CLIP-ES (Pred) depends solely on predicted class labels, aligning with the standard weakly supervised setting and offering a fair basis for comparison. This comprehensive evaluation enables us to demonstrate the robustness of our method across diverse settings.

For histology benchmarks, we compare against the histology WSOL baselines used in PixelCAM~\cite{belharbipixelcam}, including DeepMIL~\cite{deepmil}, GradCAM++~\cite{gradcampp}, LayerCAM~\cite{layercam}, SAT~\cite{SAT}, and PixelCAM~\cite{belharbipixelcam}. As \tedloc relies on a CAM module for supervision, we use PixelCAM~\cite{belharbipixelcam} by default to generate the pseudo-label CAMs. We further conduct an ablation study to evaluate the impact of the CAM module by replacing PixelCAM~\cite{belharbipixelcam} with alternative WSOL methods, including DeepMIL~\cite{deepmil}, GradCAM++~\cite{gradcampp}, and LayerCAM~\cite{layercam}.

\vspace{0.25cm}
\noindent\subsection{Comparison with State-of-the-Art Methods}

\subsubsection{Evaluation on Natural-Image Datasets}

\noindent \textbf{Quantitative results.}
The results in Tab.~\ref{finalch:tab:results_ilsvrc_cub} show that \tedloc consistently improves localization performance over previous WSOL methods on both \ilsvrc and \cubs. On \ilsvrc, our best variant (\tedloc$\langle$patch,anchor$\rangle$) attains a \maxboxacc of 77.1\%, a Top-1 \loc of 71.8\%, and a Top-5 \loc of 76.7\%. This corresponds to absolute gains of 2.1, 6.6, and 3.3 percentage points, respectively, compared to the previous best reported results from GenPrompt~\citep{zhao2023generative}. On \cubs, \tedloc reaches a \maxboxacc of 98.7\%, a Top-1 \loc of 92.0\%, and a Top-5 \loc of 97.5\%, improving on GenPrompt by 0.7, 5.0, and 1.4 percentage points, respectively. The significant improvements on both datasets highlight the capability of \tedloc in handling diverse and challenging scenarios.

In terms of classification, \tedloc also yields a substantial improvement over the distillation backbone. On \ilsvrc, our method attains a Top-1 \classification accuracy of 89.9\%, whereas the underlying CLIP model reaches 67.3\%. On \cubs, \tedloc achieves 93.0\% Top-1 \classification accuracy compared to 46.4\% for CLIP. Thus, a single network provides improved performance in both localization and classification, supporting the hypothesis that transferring text embeddings to patch representations yields a shared representation that is discriminative for both tasks.

To show the efficacy of our proposed method, we compare its model complexity and inference speed with GenPrompt~\cite{zhao2023generative} (Tab.\ref{finalch:tab:complexity-analysis}). Although GenPrompt yields comparable localization performance, it introduces significant computational overhead. More specifically, GenPrompt employs a generative framework involving diffusion models and CLIP embeddings to generate \loc maps, resulting in a considerable increase in complexity due to the multiple-module pipeline and iterative nature of the diffusion process. Specifically, it comprises an EfficientNet-based classifier for label prediction (66.35 million parameters), a variational autoencoder (VAE) for latent embeddings (83.65 million parameters), and CLIP $E_T$ for discriminative and representation embeddings, having 123.83 million parameters. The diffusion model, utilizing a U-Net architecture, adds 859.5 million parameters and iterates across 100-time steps, drastically increasing computation time. Our proposed method thus emerges as a significantly more efficient alternative, achieving robust performance without the exorbitant computational cost characteristic of GenPrompt. In contrast, \tedloc is a compact model (569.67M parameters) and obviates the need for external classifiers or diffusion sampling. Furthermore, for inference time, we utilized an idle machine equipped with an NVIDIA A100 GPU. We first conducted 20 warm-up epochs with a batch size of 1, followed by 1,000 inference steps, and calculated the average inference time across these steps.\PRbreak{}
\begin{table}[!t]
\caption{Computational complexity of our proposed \tedloc against GenPrompt.}
\label{finalch:tab:complexity-analysis}
\tabfont
\begin{tabular*}{\linewidth}{@{\extracolsep{\fill}}lll@{}}
\toprule
\textbf{Methods} & \multicolumn{2}{l}{\textbf{Complexity Analysis}} \\
\cmidrule{2-3}
 & \# Parameters &  Inference Time \\
\midrule
GenPrompt~\cite{zhao2023generative}   &  1133.35M   & 272ms \\
\tedloc (ours)                        &  569.67M    & 121ms \\
\bottomrule
\end{tabular*}
\end{table}
\begin{figure*}[!t]
    \centering
    \includegraphics[width=0.505\linewidth]{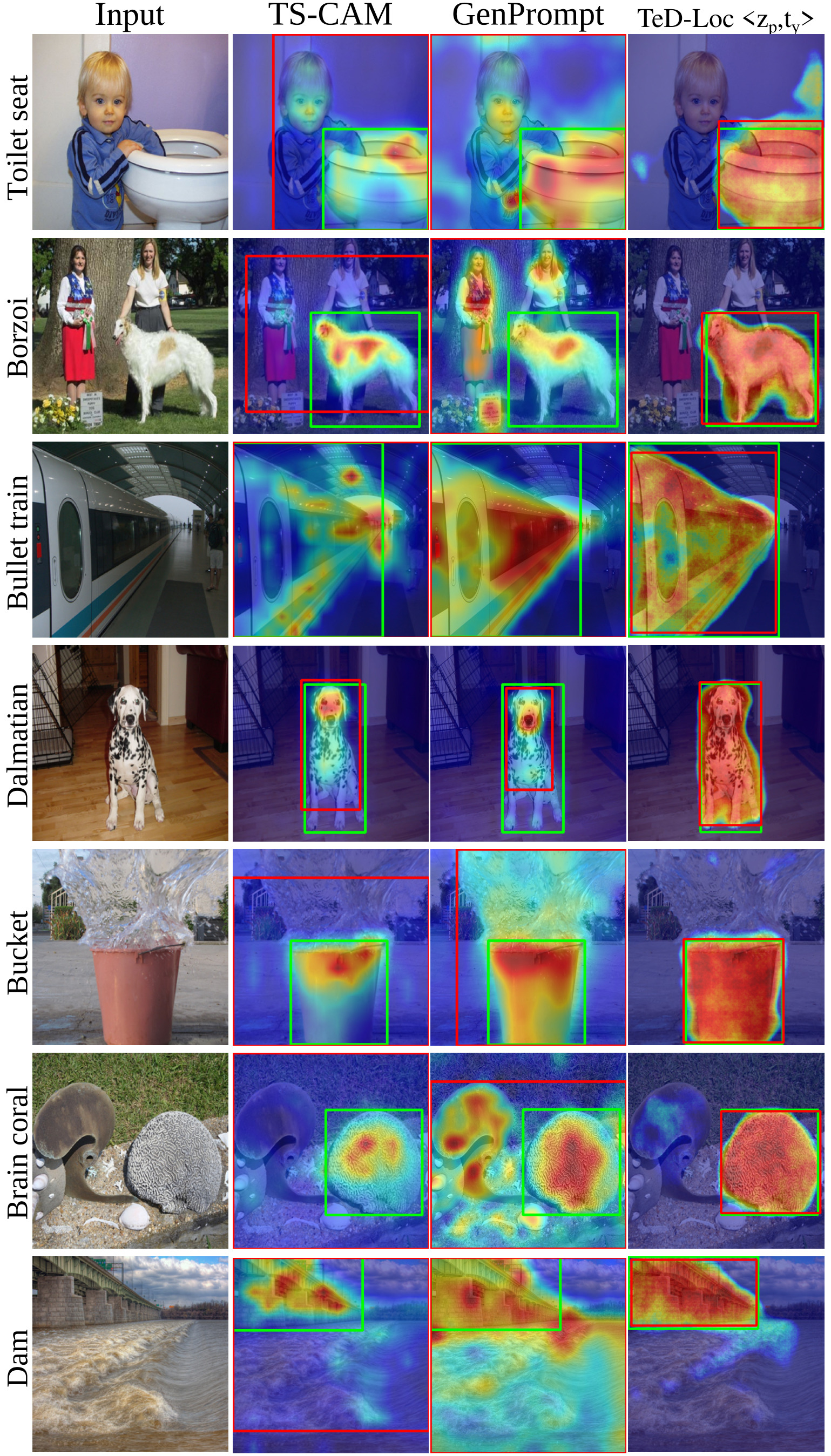}
    \caption{Qualitative comparison on \ilsvrc data. Localization maps are obtained via (patch, class) embeddings dot product: ${\langle z_p, t_y \rangle}$ where $z_p, \forall p \in \Omega$ are the patch embeddings. GenPrompt fails to localize objects in complex scenes, often because it relies on external classifiers to compute text embeddings. This can fail if the classifier makes mistakes. This dependence on class labels during inference highlights GenPrompt's vulnerability to localization errors. In contrast, \tedloc can localize objects in complex scenes. Here, \textcolor{green}{green bboxes} denote GT localization, while {red bboxes} represent predicted localizations.}
    \label{finalch:fig:main-results}
\end{figure*}
\begin{table*}[!t]
\caption{Localization (\pxap) and classification (\cl) accuracy on the \glas and \camelyonsev center-wise test datasets.}
\label{tab:accuracy-bloc-c17}
\tabfont
\setlength{\tabcolsep}{3pt}
\begin{tabular*}{\textwidth}{@{\extracolsep{\fill}}lcccccccccccc@{}}
\toprule
\textbf{Methods}
& \multicolumn{2}{l}{\textbf{\glas}}
& \multicolumn{2}{l}{\camelyonsevzero}
& \multicolumn{2}{l}{\camelyonsevone}
& \multicolumn{2}{l}{\camelyonsevtwo}
& \multicolumn{2}{l}{\camelyonsevthree}
& \multicolumn{2}{l}{\camelyonsevfour} \\
\cmidrule{2-3}\cmidrule{4-5}\cmidrule{6-7}\cmidrule{8-9}\cmidrule{10-11}\cmidrule{12-13}
& \pxap (\(\uparrow\)) & \cl (\(\uparrow\))
& \pxap (\(\uparrow\)) & \cl (\(\uparrow\))
& \pxap (\(\uparrow\)) & \cl (\(\uparrow\))
& \pxap (\(\uparrow\)) & \cl (\(\uparrow\))
& \pxap (\(\uparrow\)) & \cl (\(\uparrow\))
& \pxap (\(\uparrow\)) & \cl (\(\uparrow\)) \\
\midrule
DeepMIL~\citep{deepmil} {\emph{(icml,2018)}}      & 79.9 & \textbf{100.0} & 34.8 & 82.8 & 30.1 & 80.8 & 31.3 & 63.8 & 27.6 & 88.6 & 18.0 & 59.4 \\
GradCAM{\textit{++}}~\citep{gradcampp} {\emph{(wacv,2018)}} & 76.8 & \textbf{100.0} & 21.9 & 72.1 & 22.2 & 66.3 & 29.8 & 80.8 & 32.4 & 77.6 & 21.4 & 59.7 \\
LayerCAM~\citep{layercam} {\emph{(tip,2021)}}     & 75.1 & \textbf{100.0} & 22.8 & 72.1 & 22.6 & 66.3 & 30.1 & 80.8 & 33.1 & 77.6 & 21.8 & 66.1 \\
SAT~\citep{SAT} {\emph{(iccv,2023)}}          & 65.9 & \textbf{100.0} & 20.6 & 64.5 & 17.5 & 72.7 & 27.7 & 65.6 & 20.4 & 52.6 & 18.9 & 49.6 \\
PixelCAM~\citep{belharbipixelcam} {\emph{(midl,2025)}}     & 86.6 & \textbf{100.0} & 49.8 & 80.9 & 52.2 & 73.8 & 54.9 & 73.4 & 71.9 & 65.1 & 50.6 & 61.1 \\
\midrule
\tedloc       & \textbf{88.8} & \textbf{100.0} &  \textbf{76.2} &  \textbf{93.5} & \textbf{79.7} &  \textbf{95.3} & \textbf{71.7} &  \textbf{89.1} &  \textbf{85.1} & \textbf{94.4} &  \textbf{82.0} &  \textbf{92.9} \\
\bottomrule
\end{tabular*}
\end{table*}

\noindent \textbf{Qualitative results.} Fig.\ref{finalch:fig:main-results} presents a visual comparison of our method against state-of-the-art WSOL approaches, specifically GenPrompt~\cite{zhao2023generative} and TS-CAM~\cite{gao2021ts}, on the \ilsvrc dataset. While these methods yield competitive quantitative performance, they often struggle with accurately localizing complex objects. They tend to highlight irrelevant parts or even entirely different objects, especially in intricate scenes. GenPrompt, for instance, relies on CLIP's discriminative and representative embeddings during inference, which necessitates class labels at test time using the external classifier. While this approach aims to adaptively focus on the object of interest, it can erroneously localize incorrect objects when the classifier predicts the wrong class. This dependency on class information during inference increases the chances of mistakes, especially in critical applications where precise localization without prior class knowledge is required. In contrast, our method consistently achieves high localization accuracy by effectively capturing both discriminative and non-discriminative regions of the target object. These localization maps can produce bboxes that encompass the entire object, enhancing both localization performance and interpretability. Unlike other methods that produce low-activation regions resulting in bboxes over areas without meaningful content, our approach ensures that activations correspond closely with the actual visual appearance of the object.

Furthermore, Fig.\ref{finalch:fig:main-results} shows different failure patterns for the baseline methods. For instance, TS-CAM highlight the most salient co-occurring object rather than the target (e.g., the child rather than the toilet seat), reflecting its partial-activation bias. Also, GenPrompt fails when its external classifier mispredicts the class, since an incorrect label (from the external classifier) propagates a localization. TeD-Loc able to deal with these fail modes as the distilled patch embeddings carry semantic information, producing sharp activation maps across different scenes.

\subsubsection{Evaluation on Histology Datasets}

To assess the generalization of \tedloc beyond natural-image benchmarks, we evaluate it on two challenging histology datasets: \glas for colon cancer and \camelyonsev for breast cancer, which\PRbreak{} includes data from five centers. Unlike natural images, histology images present distinct challenges, as regions of interest often resemble the surrounding background and substantial variability may exist within the same class~\cite{guichemerre2024source, belharbipixelcam}. Consequently, WSOL methods tend to perform poorly in histology, as relevant regions are often difficult to distinguish and localize using image-level supervision alone. To address these challenges, recent methods such as PixelCAM~\cite{belharbipixelcam} have been specifically developed to improve weakly supervised localization in this setting. As such, histology provides a rigorous benchmark for evaluating the robustness and generalizability of WSOL approaches.

\noindent {\textbf{Quantitative results.}} Tab.~\ref{tab:accuracy-bloc-c17} shows that \tedloc consistently outperforms all WSOL models, including the recent PixelCAM~\cite{belharbipixelcam}, on both localization and classification. On \glas, classification is saturated for all methods (100.0\%), so the meaningful comparison is localization, where \tedloc reaches 88.8\% \pxap. On \camelyonsev, the gains are larger and consistent across all test centers. Compared to the recent method PixelCAM~\citep{belharbipixelcam}, \tedloc improves localization across all centers by +26.4\%, +27.5\%, +16.8\%, +13.2\%, and +31.4\%, respectively. Averaged across the \camelyonsev centers, this corresponds to an improvement of +23.1\% in localization and +22.2\% in classification over PixelCAM~\citep{belharbipixelcam}. These gains indicate that the proposed method not only produces better activation maps, but also leads to more reliable classification under cross-center variability in histology.

To assess whether the gains of \tedloc are driven by the pretrained VLM, we compare zero-shot performance when using either the pathology-specialized CONCH initialization or an initialization pretrained on natural scene images. As we can observe in Tab.~\ref{tab:tedloc-conch-zeroshot}, using the pretrained model CONCH~\cite{conch} did not impact the classification accuracy, with similar results on \glas and \camelyonsev. For localization, CONCH provides slight improvements overall, with the largest gain observed on \glas (+4.4\% \pxap). Regarding \camelyonsev, the differences remain small and inconsistent across centers. These results highlight that the observed performance of \tedloc is mainly due to the proposed method rather than being explained by the encoder pretraining.

\noindent {\textbf{Qualitative results.} Fig.~\ref{fig:histology-wsol} presents a visual comparison of \tedloc against standard WSOL baselines on the \camelyonsevfour \emph{cancer} class. As shown in the figure, baseline CAM-based methods often activate irrelevant regions or only partial ROIs. \tedloc produces sharper and less noisy localization, reducing the false positives while preserving the relevant ROIs of the tissues.\PRbreak{}
\begin{table*}[!t]
\caption{Zero-shot localization (\pxap) and classification (\cl) accuracies of \tedloc with and without CONCH initialization on \glas and \camelyonsev center-wise test sets.}
\label{tab:tedloc-conch-zeroshot}
\tabfont
\setlength{\tabcolsep}{3pt}
\begin{tabular*}{\textwidth}{@{\extracolsep{\fill}}lcccccccccccc@{}}
\toprule
\textbf{Methods}
& \multicolumn{2}{l}{\textbf{\glas}}
& \multicolumn{2}{l}{\camelyonsevzero}
& \multicolumn{2}{l}{\camelyonsevone}
& \multicolumn{2}{l}{\camelyonsevtwo}
& \multicolumn{2}{l}{\camelyonsevthree}
& \multicolumn{2}{l}{\camelyonsevfour} \\
\cmidrule{2-3}\cmidrule{4-5}\cmidrule{6-7}\cmidrule{8-9}\cmidrule{10-11}\cmidrule{12-13}
& \pxap\ (\(\uparrow\)) & \cl\ (\(\uparrow\))
& \pxap\ (\(\uparrow\)) & \cl\ (\(\uparrow\))
& \pxap\ (\(\uparrow\)) & \cl\ (\(\uparrow\))
& \pxap\ (\(\uparrow\)) & \cl\ (\(\uparrow\))
& \pxap\ (\(\uparrow\)) & \cl\ (\(\uparrow\))
& \pxap\ (\(\uparrow\)) & \cl\ (\(\uparrow\)) \\
\midrule
\tedloc wo/ CONCH~\cite{conch} & 49.3 & 53.7 & 16.1 & 50.0 & 14.7 & 50.0 & 27.2 & 50.0 & 20.9 & 50.0 & 19.2 & 50.0 \\
\tedloc\ w/ CONCH~\cite{conch} & 53.5 & 53.7 & 16.7 & 50.0 & 14.8 & 50.0 & 28.0 & 50.0 & 19.8 & 50.0 & 19.4 & 50.0 \\
\bottomrule
\end{tabular*}
\end{table*}
\begin{figure*}[!t]
    \centering
    \includegraphics[width=0.68\linewidth]{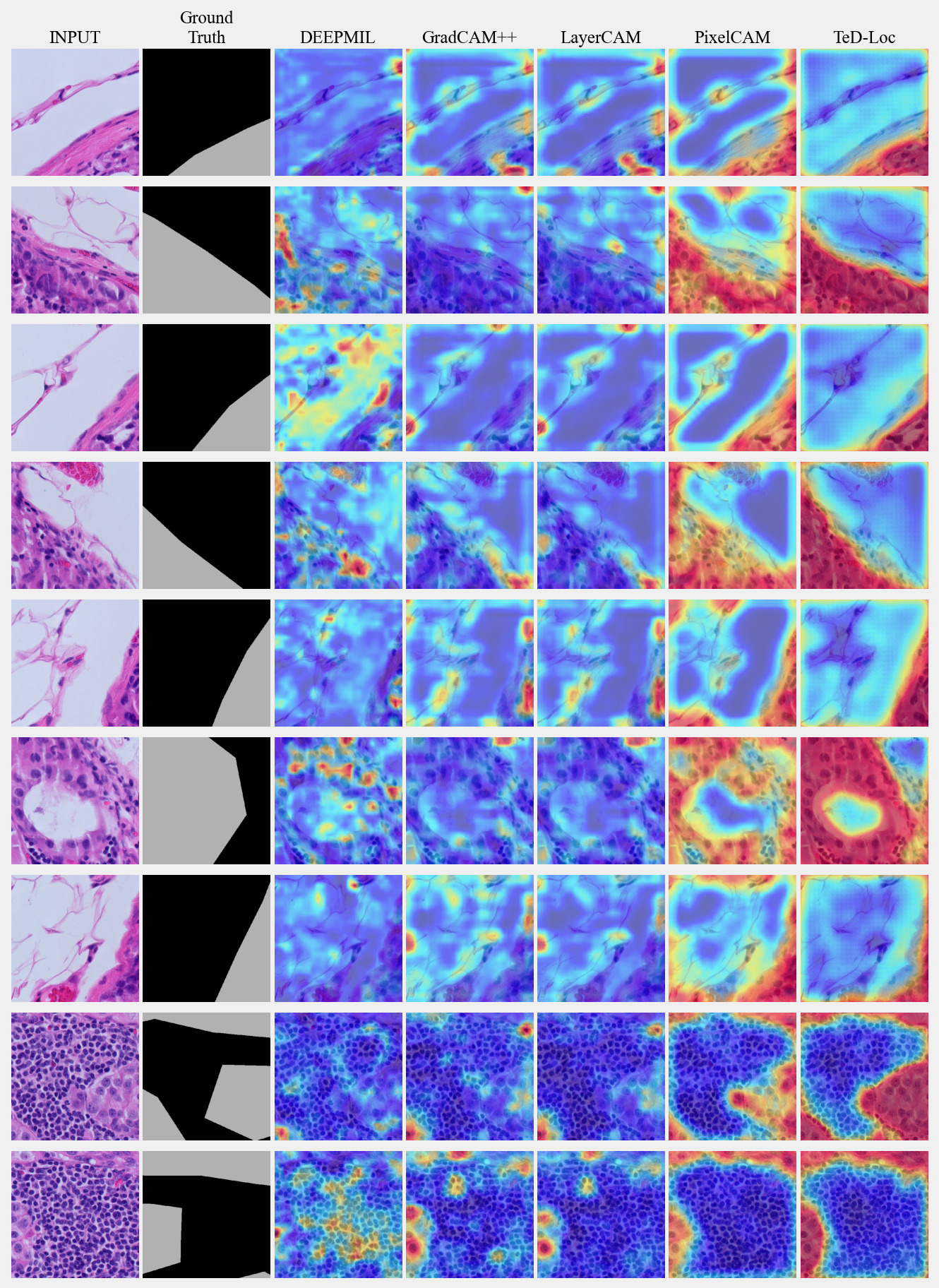}
    \caption{Visualization of localization map obtained via \tedloc as compared to standard WSOL models on the \camelyonsevfour dataset for the class cancer.}
    \label{fig:histology-wsol}
\end{figure*}
}

\subsubsection{Ablations.}\label{sec:ablations}
\noindent \textbf{Impact of losses.} Tab.\ref{finalch:tab:abla_study_cub} shows that the combination of our proposed loss functions: knowledge-distillation loss ($\mathbf{L}_{\text{KD}}$), patch classifier loss\PRbreak{} ($\mathbf{P}_{\text{CL}}$) and global image classification loss ($\mathbf{I}_{\text{CL}}$), is essential for achieving state-of-the-art localization performance. Using only the main loss $\mathbf{L}_{\text{KD}}$, our method yields a \maxboxacc of 62.3\%, \maxboxacc of 62.3\%, as it focuses on learning FG embeddings without yielding explicit\PRbreak{}
\begin{figure*}[!t]
    \centering
    \includegraphics[width=0.58\linewidth]{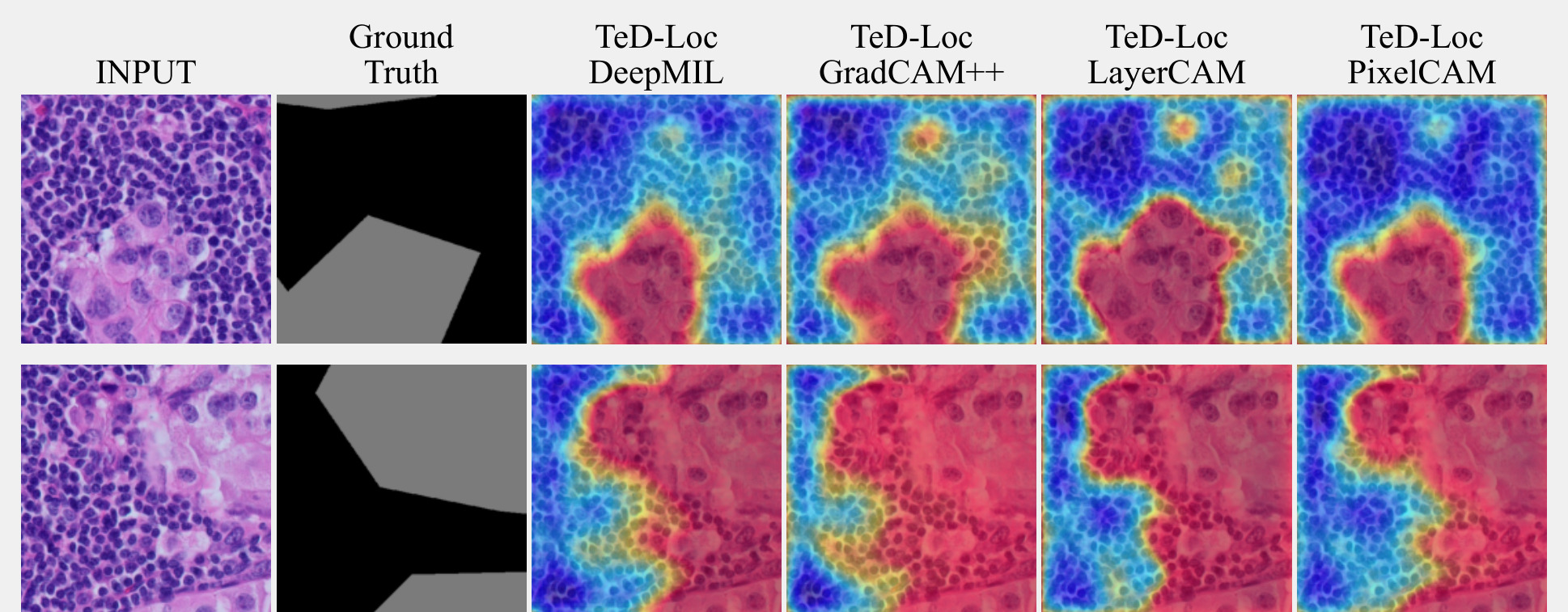}
    \caption{Visualization of localization map obtained via \tedloc on the \camelyonsevfour dataset for the class cancer with different CAM modules (DeepMil~\cite{deepmil}, GradCAM++~\cite{gradcampp}, LayerCAM~\cite{layercam}, PixelCAM~\cite{belharbipixelcam}).}
    \label{fig:visu-cam-module}
\end{figure*}
 localization, thereby explaining its limited performance. Adding patch classifier loss significantly improves accuracy to 95.4\% by enhancing FG/BG separation. Finally, incorporating image class loss boosts the performance to 98.7\%, emphasizing the importance of discriminative learning to distinguish between correct and incorrect class alignments. This demonstrates the effect of integrating these losses to achieve highly accurate weakly supervised object localization.

\noindent \textbf{Hyperparameters sensitivy.} Tab~\ref{tab:tedloc-cam-module} analyzes the sensitivity of \tedloc to the loss weights $\lambda_1$, $\lambda_2$, $\lambda_3$, which control the contribution of\PRbreak{} the knowledge-distillation, patch classification, and image classification losses, respectively. Results on \glas show that \tedloc performs best under a balanced weighting of these objectives. In particular, the best performances are obtained if all lambdas are $\le$ 1. Larger values generally lead to a decline in \pxap.

\noindent \textbf{Impact of different CAM modules on \tedloc performance.} Tab~\ref{tab:tedloc-cam-module} analyzes the impact of the CAM pseudo-labels source on \camelyonsevfour. While the choice of CAM slightly affects performance, all different modules achieve strong results, with localization ranging from 78.3\PRbreak{}
\begin{table}[!t]
\caption{Ablation study on the \cubs dataset showing the impact of different loss combinations on \maxboxacc performance.}
\label{finalch:tab:abla_study_cub}
\tabfont
\begin{tabular*}{\linewidth}{@{\extracolsep{\fill}}lr@{}}
\toprule
\textbf{Losses} & \textbf{CUB (\maxboxacc)}\\
\midrule
$\mathbf{L}_{\text{KD}}$ & 62.3 \\
$\mathbf{L}_{\text{KD}}$+$\mathbf{P}_{\text{CL}}$ & 95.4 \\
$\mathbf{L}_{\text{KD}}$+$\mathbf{P}_{\text{CL}}$+$\mathbf{I}_{\text{CL}}$ & \textbf{98.7} \\
\bottomrule
\end{tabular*}
\end{table}
\begin{table}[!t]
\caption{Ablation study of the loss weights on \glas. Columns report the values tested for \(\lambda_1\), \(\lambda_2\), and \(\lambda_3\).}
\label{tab:ablation-lambdas-landscape}
\tabfont
\setlength{\tabcolsep}{2.5pt}
\begin{tabular*}{\linewidth}{@{\extracolsep{\fill}}lcccccccccccc@{}}
\toprule
\textbf{Dataset}
& \multicolumn{4}{l}{\textbf{\(\lambda_1\)}}
& \multicolumn{4}{l}{\(\lambda_2\)}
& \multicolumn{4}{l}{\(\lambda_3\)} \\
\cmidrule{2-5}\cmidrule{6-9}\cmidrule{10-13}
& 0.5 & 1.0 & 1.5 & 2.0
& 0.5 & 1.0 & 1.5 & 2.0
& 0.5 & 1.0 & 1.5 & 2.0 \\
\midrule
\glas            & 89.1 & 88.8 & 78.7 & 78.2 & 78.1 & 88.8 & 78.6 & 78.6 & 78.3 & 88.8 & 86.4  &  78.8 \\
\bottomrule
\end{tabular*}
\end{table}
\begin{table}[!t]
\caption{Impact of different CAM-based pseudo-labels on \tedloc performance. Localization (\pxap) and classification (\cl) accuracies on test set of \camelyonsevfour dataset.}
\label{tab:tedloc-cam-module}
\tabfont
\begin{tabular*}{\linewidth}{@{\extracolsep{\fill}}lcc@{}}
\toprule
\textbf{Methods} & \multicolumn{2}{l}{\textbf{\camelyonsevfour}} \\
\cmidrule{2-3}
& \pxap\ (\(\uparrow\)) & \cl\ (\(\uparrow\)) \\
\midrule
DeepMIL~\cite{deepmil}      & 18.0 & 59.4 \\
\tedloc w/ DeepMIL      & \textbf{78.3} & \textbf{89.3} \\
\midrule
GradCAM{\textit{++}}~\cite{gradcampp} & 21.4 & 59.7 \\
\tedloc w/ GradCAM{\textit{++}} & \textbf{81.1} & \textbf{94.3} \\
\midrule
LayerCAM~\cite{layercam}     & 21.8 & 66.1 \\
\tedloc w/ LayerCAM     & \textbf{80.4} & \textbf{93.9} \\
\midrule
PixelCAM~\cite{belharbipixelcam}     & 50.6 & 61.1 \\
\tedloc w/ PixelCAM     & \textbf{82.0} & \textbf{92.9} \\
\bottomrule
\end{tabular*}
\end{table}
\begin{figure*}[!t]
    \centering
    \includegraphics[width=0.64\linewidth]{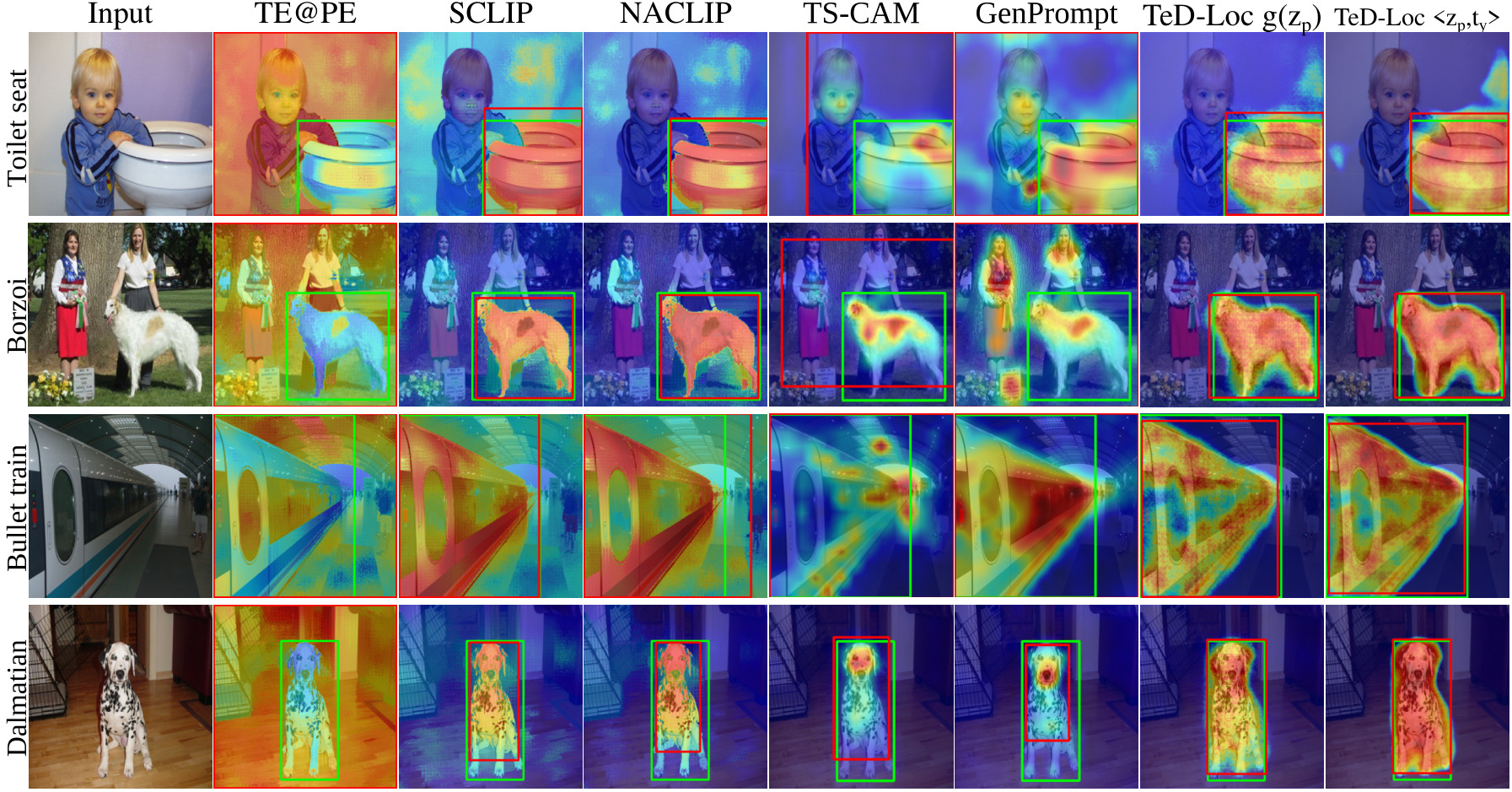}
    \caption{Visualization of localization map obtained via $g(z_p)$ as compared to (patch, class) embeddings dot product: ${\langle z_p, t_y \rangle}$ where $z_p, \forall p \in \Omega$ is the patch embeddings, and $y$ is the true image class over different variants of \clip, and our method. TE@PE is the vanilla \clip~\cite{radford2021learning} where TE is the text embedding, and PE is the patch embedding.}
    \label{fig:patch-level-score}
\end{figure*}
\begin{figure*}[!t]
    \centering
    \includegraphics[width=0.63\linewidth]{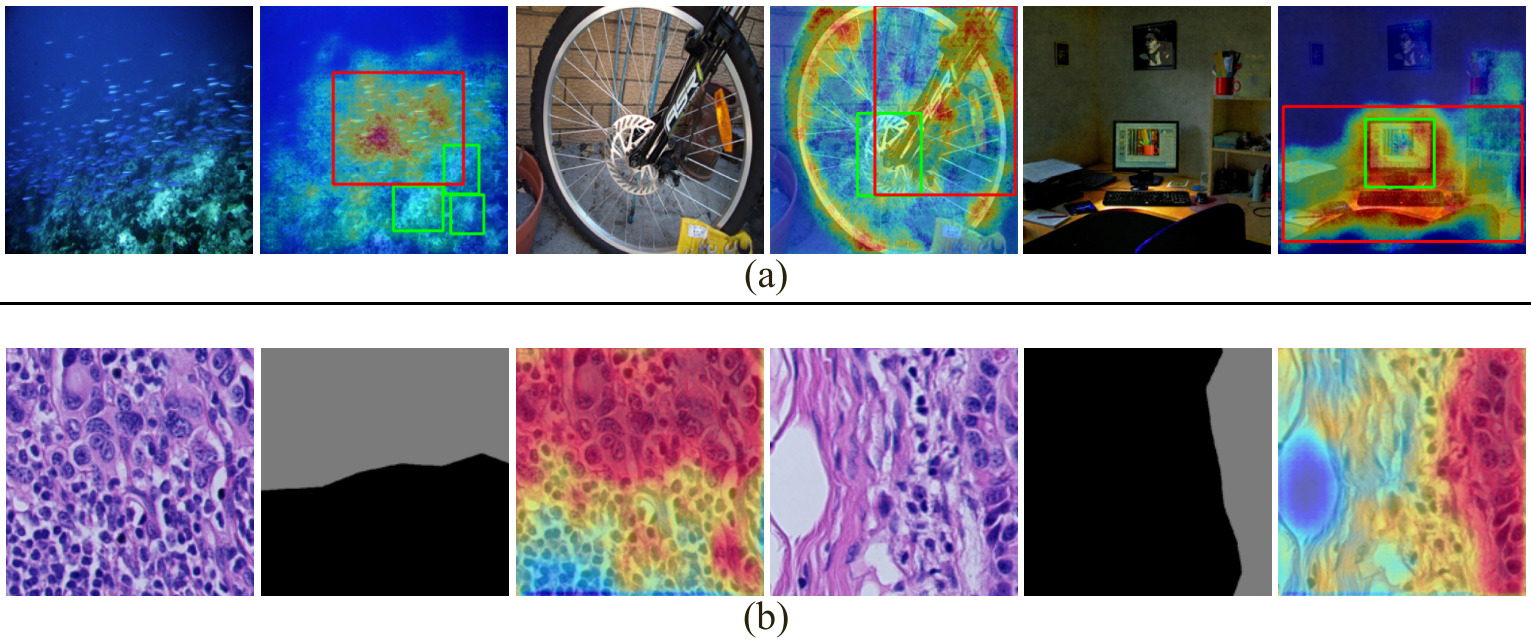}
    \caption{Failure cases of \tedloc in localizing ROIs. Row (a) shows 3 examples from natural images (\ilsvrc) where \gt annotations are bboxes and localization is evaluated via \maxboxacc. Row (b) shows 3 examples from histology images (\camelyonsev), where \gt annotations are pixel-level segmentation masks, and localization is evaluated via \pxap}
    \label{fig:failuercaes}
\end{figure*}
 to 82.0 \pxap and classification ranging from 89.3 to 94.3. These performances are higher than those of the corresponding standard WSOL models reported in Tab~\ref{tab:accuracy-bloc-c17}, whose localization performance ranges from 18.0 to 50.6 \pxap on the center \camelyonsevfour. Also, Fig.~\ref{fig:visu-cam-module} shows that \tedloc produces qualitatively similar localization maps across different CAM modules. Although stronger CAM modules can yield slightly cleaner boundaries and fewer false-positive activations, the overall gain appears largely independent of the specific CAM module used.

\vspace{-1pt}

\subsection{Patch-level Localization with Text Anchors vs Localization from $g(z_p)$}
In the standard \clip model~\cite{radford2021learning}, text class embedding is not necessarily correlated with the local vision patch embedding. To show this, we conduct the following experiment: consider the text embedding of the image class label $y$: $t_y$. Then, to localize this class object within the image, we perform a dot product across all patch embeddings: ${\langle z_p, t_y \rangle}$ where $z_p, \forall p \in \Omega$. High scores at location $p$ should indicate the\PRbreak{}
\begin{table*}[!t]
    \centering
    \begin{minipage}{0.8\textwidth}
    \caption{Localization performance of maps obtained via $g(z_p)$ and (patch, class) embeddings dot product: ${\langle z_p, t_y \rangle}$ where $z_p, \forall p \in \Omega$ is the patch embeddings, and $y$ is the true image class. We report localization performance (\textbf{\maxboxacc},\toponebold \locbold, \topfivebold \locbold) using different variants of \clip and our method.}
    \label{finalch:tab:clip-perf}
    \tabfont
    \begin{tabular*}{\linewidth}{@{\extracolsep{\fill}}lcccccc@{}}
    \toprule
     & \multicolumn{3}{l}{\cubsbold} & \multicolumn{3}{l}{\ilsvrcbold} \\
    \cmidrule{2-4}\cmidrule{5-7}
     & \textbf{\maxboxacc} & \toponebold \locbold & \topfivebold \locbold & \textbf{\maxboxacc} & \toponebold \locbold & \topfivebold \locbold \\
    \midrule
    Vanilla \clip~\cite{radford2021learning} & 18.8 & 8.9 & 15.2 & 41.1 & 26.6 & 37.0 \\
    SCLIP (eccv'24)~\cite{wang2023sclip}  & 85.8 & 14.4 & 37.1 & 70.4 & 33.9 & 55.1 \\
    NACLIP (wacv'25)~\cite{sina25wacv} & 80.8 & 12.0 & 32.3 & 71.7 & 28.6 & 49.4 \\
    \midrule
    \tedloc w/ $g(z_p)$ & \textbf{75.6} & \textbf{70.0} & \textbf{75.1} & \textbf{98.7} & \textbf{91.7} & \textbf{97.6} \\
    \tedloc w/ ${\langle z_p, t_y \rangle}$ & \textbf{77.2} & \textbf{71.8} & \textbf{76.7} & \textbf{98.7} & \textbf{92.0} & \textbf{97.5} \\
    \bottomrule
    \end{tabular*}
    \end{minipage}
\end{table*}
 high likelihood of the object $y$ presence at this location. The obtained score map is then considered as a CAM. We perform this experiment over three variants of \clip model: Vanilla \clip~\cite{radford2021learning}, SCLIP~\cite{wang2023sclip}, and NACLIP~\cite{sina25wacv}, in addition to our method. The obtained results are presented in Tab.\ref{finalch:tab:clip-perf}. Vanilla \clip yielded poor results, confirming that class text embeddings are not necessarily correlated with the local patch embeddings, making them less useful for this task. This justifies using a gradient-based method over the dot product score between the global image embedding and a class text embedding in CLIP-ES~\cite{lin23}. However, we notice a greater improvement in localization for the more recent \clip variants SCLIP~\cite{wang2023sclip}, and NACLIP~\cite{sina25wacv}. However, their \toponebold, and \topfivebold  \locbold are still low, indicating poor classification performance. On the other hand, our method achieves the highest performance over the three metrics, indicating better localization and classification scores over both datasets. This is the result of our text-to-patch distillation, which ensures that local patch embeddings are correlated with the class text embedding, allowing direct localization based on the text embedding. This equips our method with a secondary localization approach and the patch FG/BG classifier $g$. This second localization approach yielded relatively better performance than when using $g$. In addition to this, we evaluate the localization map extracted from module $g(z_p)$ of our method, which performs slightly lower than the results obtained via ${\langle z_p, t_y \rangle}$ as shown in Tab.~\ref{finalch:tab:clip-perf}. Furthermore, Fig.~\ref{fig:patch-level-score} visualizes this localization strategy of different \clip variants and our method.

\noindent \textbf{Impact of Orthogonalization on Performance.} Table~\ref{finalch:tab:res-anchors} shows the impact of orthogonalization of class text embedding in our method for both tasks: classification and localization (obtained via $g(z_p)$). These results suggest that original class text embeddings of \clip~\cite{radford2021learning} are not adequate to perform discriminative learning, as these embeddings overlap, as shown visually in the main paper (Fig.\ref{finalch:fig:orth-embds}). However, their orthogonalization allows better separation of these embeddings, making them more suitable for classification tasks. This also positively affects localization as well in our method, since both tasks are strongly related by design.

\subsection{Failures cases}

\tedloc provides clear advantages and significantly improves localization performance over standard WSOL models, but it still fails in some challenging cases for both natural and medical images. For natural images, Fig.~\ref{fig:failuercaes}(a) illustrates several typical failure modes. In the ``coral reef'' example, the presence of fish within the reef scene appears to mislead the model into partially associating the target class with the fish, which results in activations over both the fish and the reef rather than a more precise localization of the coral region. In the ''bicycle tire'' example, the activation extends beyond the target part and covers a broader portion of the wheel, suggesting limited spatial precision for fine-grained categories. In the ''desktop computer'' example, the model\PRbreak{}
\begin{table}[!t]
    \caption{Impact of class text embeddings (text anchors) orthogonalization over localization and classification performance in our method over \cubsbold dataset.}
    \label{finalch:tab:res-anchors}
    \tabfont
    \begin{tabular*}{\linewidth}{@{\extracolsep{\fill}}p{0.42\linewidth}cccc@{}}
    \toprule
    \tedlocbold & \multicolumn{4}{l}{\cubsbold} \\
    \midrule
    w/o orthogonalization, with default anchors and $g(z_p)$  & 97.7 & 56.0 & 54.8 & 85.8 \\
    w/ orthogonalization, with $g(z_p)$ & \textbf{98.7} & \textbf{93.0} & \textbf{91.7} & \textbf{97.6} \\
    \bottomrule
    \end{tabular*}
\end{table}
 also responds to the surrounding desk area, indicating that contextual cues may dominate when the object occupies only a limited portion of the image. More generally, these examples show that, although \tedloc identifies semantically relevant regions, the predicted localization can remain spatially coarse and may over-extend beyond the true object boundaries.\PRbreak{}

A similar limitation is observed in histology images, as shown in Fig.~\ref{fig:failuercaes}(b). In these examples, \tedloc correctly assigns high confidence to diagnostically relevant tissue regions, but the activation map may still spread into adjacent background. This suggests that the method captures the most informative regions, yet does not always delineate the ROI sharply. Such behavior is consistent with the intrinsic difficulty of histology WSOL, where boundaries between relevant and non-relevant tissue are often gradual, local morphology can be highly heterogeneous, and the amount of annotated training data is limited~\cite{guichemerre2026adaptwsol}. These observations indicate that, while text-guided alignment improves semantic localization, additional spatial regularization or refinement strategies may be required to achieve more precise boundary delineation in challenging settings characterized by limited data and strong foreground-background similarity.\PRbreak{}

\section{Conclusion}
We have introduced \tedloc, a novel WSOL approach that integrates textual and visual modalities by transferring knowledge directly from CLIP's text embeddings to our patch embedding module. This alignment enables our model to localize objects at the patch-level while simultaneously performing global image classification. Our results show that language-guided visual alignment can effectively benefit WSOL tasks and further suggest that stronger localization can directly benefit image-level recognition. Additionally, the proposed QR-based text embedding orthogonalization improves discriminability for semantically similar classes. Experiments on \cubs, \ilsvrc, and two histopathology benchmarks (\glas and \camelyonsev) show consistent improvements over baseline methods. For instance, our method improves up to 31\% \pxap on histopathology datasets, while offering more efficient inference than GenPrompt. Regarding limitations, although \tedloc improves localization performance, the qualitative results show that it can still fail in challenging cases, either by missing parts of the relevant ROI or by producing over-activations. These failures suggest that semantic text-guided alignment alone is not always sufficient to guarantee precise spatial localization, and that additional spatial regularization or refinement strategies could further improve localization quality. While the current study focuses on single-object localization benchmarks, the class-conditioned design of \tedloc suggests that it could naturally extend to multi-class or multi-instance settings. However, this potential remains to be validated experimentally. Finally, the behavior of CLIP-based text guidance in highly specialized domains, such as remote sensing or industrial inspection, beyond those studied here remains to be investigated. We believe \tedloc opens new directions for WSOL by distilling class-level semantic knowledge from the CLIP text encoder into patch representations using only image-level supervision. The text distillation paradigm could be plugged into the backbones. Future work could explore extending \tedloc to open-vocabulary localization settings and to multi-label scenarios, and investigate domain adaptation to improve generalization across histopathology datasets.

\renewcommand{\appendixname}{}

\renewcommand{\thesection}{\Alph{section}}
\renewcommand{\thesubsection}{\thesection.\arabic{subsection}}

\bibliographystyle{elsarticle-num}
\bibliography{egbib}

\end{document}

%% file: macros.tex
\usepackage{xcolor}
\newcommand\tedloc{\texttt{TeD-Loc}\xspace}
\newcommand\tedlocbold{{\fontfamily{lmtt}\fontseries{b}\selectfont TeD-Loc}\xspace}
\newcommand\cubs{\texttt{CUB}\xspace}
\newcommand\ilsvrc{\texttt{ILSVRC}\xspace}
\newcommand\ilsvrcvtwo{\texttt{ILSVRC-V2}\xspace}
\newcommand\glas{\texttt{GlaS}\xspace}
\newcommand\camelyonsev{\texttt{CAMELYON17}\xspace}
\newcommand\camelyonsevzero{\texttt{C17-0}\xspace}
\newcommand\camelyonsevone{\texttt{C17-1}\xspace}
\newcommand\camelyonsevtwo{\texttt{C17-2}\xspace}
\newcommand\camelyonsevthree{\texttt{C17-3}\xspace}
\newcommand\camelyonsevfour{\texttt{C17-4}\xspace}
\newcommand\ilsvrcbold{\boldcourier{\fontfamily{lmtt}\fontseries{b}\selectfont ILSVRC}\xspace}
\newcommand\cubsbold{\boldcourier{\fontfamily{lmtt}\fontseries{b}\selectfont CUB}\xspace}
\newcommand\maxboxacc{\texttt{MaxBoxAcc}\xspace}
\newcommand\pxap{\texttt{PxAP}\xspace}
\newcommand\classification{\texttt{CL}\xspace}
\newcommand\topone{\texttt{Top-1}\xspace}
\newcommand\topfive{\texttt{Top-5}\xspace}
\newcommand\iou{\texttt{IoU}\xspace}
\newcommand{\fakeslant}[2][0.25em]{\makebox[2.5pt][l]{#2}\kern-#1#2}
\newcommand{\boldcourier}[1]{\textbf{\texttt{\fakeslant{#1}}}}
\newcommand\toponebold{\boldcourier{\fontfamily{lmtt}\fontseries{b}\selectfont Top-1}\xspace}
\newcommand\topfivebold{\boldcourier{\fontfamily{lmtt}\fontseries{b}\selectfont Top-5}\xspace}
\newcommand\locbold{\boldcourier{\fontfamily{lmtt}\fontseries{b}\selectfont\xspace Loc}\xspace}
\newcommand\loc{\texttt{Loc}\xspace}
\newcommand\cl{\texttt{CL}\xspace}
\newcommand\gt{\texttt{GT}\xspace}
\newcommand\clip{\texttt{CLIP}\xspace}
\newcommand\gtknown{\texttt{GT-Known}\xspace}
\newcommand\corloc{\texttt{CorLoc}\xspace}